\documentclass[a4paper, 10pt, conference]{ieeeconf}  
\usepackage{FG2024}
\usepackage{amssymb}
\usepackage{amsmath}
\usepackage{graphicx}
\usepackage{multirow}
\usepackage{xcolor,colortbl}
\usepackage{arydshln}
\usepackage{caption}
\usepackage{url}
\usepackage{hyperref}
\hypersetup{
    colorlinks=true,
    linkcolor=blue,
    citecolor=red,
    filecolor=magenta,      
    urlcolor=cyan,
}

\FGfinalcopy

\definecolor{p_tab_best}{RGB}{255, 153, 153}
\definecolor{p_tab_sbest}{RGB}{255, 204, 153}
\def\best{\bf \cellcolor{p_tab_best}}
\def\secbest{\cellcolor{p_tab_sbest} }

\IEEEoverridecommandlockouts                             

\newcommand{\norm}[1]{\left\lVert#1\right\rVert}
\newcommand\blfootnote[1]{%
  \begingroup
  \renewcommand\thefootnote{}\footnote{#1}%
  \addtocounter{footnote}{-1}%
  \endgroup
}

\def\FGPaperID{152} 

\title{\LARGE \bf
Diversity-Aware Sign Language Production through a Pose Encoding Variational Autoencoder
}

\author{\parbox{16cm}{\centering
    {\large Mohamed Ilyes Lakhal, Richard Bowden}\\
    {\normalsize
    CVSSP, University of Surrey, Guildford, United Kingdom}\\
    {\normalsize \texttt {\{m.lakhal, r.bowden\}@surrey.ac.uk}}}%
}

\begin{document}

\ifFGfinal
\thispagestyle{empty}
\pagestyle{empty}
\else
\author{Anonymous FG2024 submission\\ Paper ID \FGPaperID \\
}
\pagestyle{plain}
\fi

\twocolumn[{%
\renewcommand\twocolumn[1][]{#1}%
\maketitle
\begin{center}
    \centering
    \includegraphics[width=.9\textwidth]{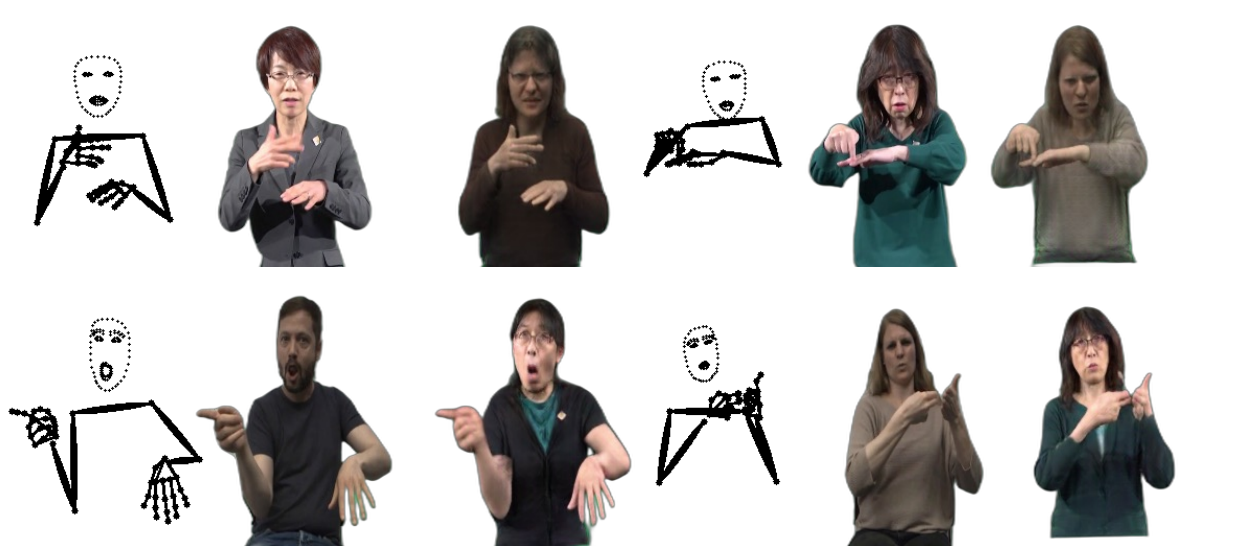}
    \captionof{figure}{\textbf{Diversity-Aware Sign Language Production}. Given an input image of a signer, we would like to synthesize a novel (unseen) image of another signer given an attribute (\textit{e.g.}, ethnicity) and a corresponding pose. KEY -- top-row: given
a Japanese signer, we produce a Swiss signer with the same pose (Japanese $\to$ Swiss); bottom-row: Swiss $\to$ Japanese.}
    \label{fig:soa_ethnicity}
\end{center}%
}]

\begin{abstract}
This paper addresses the problem of diversity-aware sign language production, where we want to give an image (or sequence) of a signer and produce another image with the same pose but different attributes (\textit{e.g.} gender, skin color). To this end, we extend the variational inference paradigm to include information about the pose and the conditioning of the attributes. This formulation improves the quality of the synthesised images. The generator framework is presented as a UNet architecture to ensure spatial preservation of the input pose, and we include the visual features from the variational inference to maintain control over appearance and style. We generate each body part with a separate decoder. This architecture allows the generator to deliver better overall results. Experiments on the SMILE II dataset show that the proposed model performs quantitatively better than state-of-the-art baselines regarding diversity, per-pixel image quality, and pose estimation. Quantitatively, it faithfully reproduces non-manual features for signers.
\end{abstract}

\section{Introduction}
In this paper, we present an approach for diversity-aware sign language production, in which we aim to synthesize a signer from a $2$D pose sequence and a defined set of attributes (\textit{e.g.}, gender, skin color). To our knowledge, this is the first to tackle this challenging problem. To do so, we introduce a pose-encoding VAE (PE-VAE), a variational inference formulation, to learn the distribution of images of signers with different attributes. In particular, we explicitly encode the pose information as part of the KL divergence loss. We combine the latent features from variational inference with attribute informatio. By incorporating the pose into the KL loss, it helps to ensure that the style code does not depend on the pose and learns a better visual feature representation.
We propose PENet, a UNet encoder-decoder model with PE-VAE as the visual feature sampler. We generate each body part separately to produce higher visual-quality images. Additionally, we add an edge loss to penalize the synthesis in edge space as we found that this helps the model to produce crisp results.

 To summarise, the contribution of this paper is three-fold:
\begin{itemize}
    \item  We propose a new formulation to the Sign Language Production to account for anonymization through pre-defined attributes.
    \item  We extend the Variational inference paradigm to learn pose-agnostic features.
    \item  We implement PE-VAE within the UNet generator where the pose is conditioned with the pose agnostic feature and an attribute to synthesize a signer.
\end{itemize}
The rest of the paper is presented as follows: Sec.~\ref{sec:back} reviews works related to image synthesis and sign language production. Sec.~\ref{sec:met} introduces our proposed PENet and the training losses. Sec.~\ref{sec:exp} overviews our experimental protocol and discusses our results against state-of-the-art baselines. Finally, Sec.~\ref{sec:conc} summarises the contributions and results of this paper.


\section{Background} \label{sec:back}
\textbf{Sign language production.} Sign Language Production (SLP) ~\cite{Razieh_2021_Arxiv} automatically translates a spoken language sequence (Text~\cite{Stoll_2020_IJCV}, or Speech~\cite{Kapoor20214391}) into a sign language sequence. The sequence can either be represented as an intermediate representation (also called Gloss~\cite{moryossef2023baseline}) or directly in a human-readable format (\textit{e.g.}, human pose~\cite{hwang2021non,Saunders_2020_ECCV}). Another way to produce Sign Language is to represent the sign through graphical avatars~\cite{Kipp_2011_IVA}. However, such a solution is less favorable for the deaf community, as small inconsistencies can lead to misleading signs~\cite{rosalee-wolfe-etal-2021-myth}. Therefore, photorealistic SLP~\cite{Saunders_ArXiv_2020} was proposed as an alternative to graphical avatars. These models use neural network encoder-decoder architectures to synthesize a sign sequence using identity-specific conditioning~\cite{Saunders_2022_CVPR,Saunders_ArXiv_2020} or anonymize the signer using variational inference~\cite{Saunders_2021_FG,Silveira_2022_SIBGRAPI}.

\textbf{Neural image synthesis.} Neural Radiance Fields (NeRF) ~\cite{mildenhall2020nerf} predict a scene's RGB value and volume density based on the position of a 3D point and its viewing direction. Human-NeRF~\cite{weng_humannerf_2022_cvpr,peng2021animatable,peng2021neural} uses a human 3D mesh template (usually SMPL~\cite{SMPL_2015_TOG}) which is deformed with a given body pose and rendered to the target viewpoint using Volume Rendering~\cite{Lichtenbelt_1998}. Despite the excellent performance of this class of methods, an incorrect estimation (\textit{e.g.} SMPL) can lead to poor rendering.

Diffusion models~\cite{ho2020denoising} are generative models that extend Variational Autoencoders (VAE)~\cite{Diederik_2019_Arxiv} to model a complex distribution through a series of Gaussian distributions using a Markovian assumption. We categorise the models into Denoising Diffusion Probabilistic Models~\cite{ho2020denoising}, where the goal is to synthesize the target in pixel space, and Latent Diffusion Models (LDMs), where the diffusion process takes place in the latent space extracted from a pre-trained model to reduce computational power~\cite{Rombach_2022_CVPR}. In the context of SLP, DiffSLVA~\cite{xia2023diffslva} takes a video and anonymises it with a text prompt. However, providing the edges from the input video provides enough identity information.

\textbf{Pose-guided human image synthesis.} This problem uses an image and a 2D human pose and synthesises an image with the target pose. The pose could be combined with the image in a single encoder~\cite{Ma_2017_NIPS} or separate encoders~\cite{Knoche_2020_CVPRW}. However, it has been shown that separate encoders are helpful for the overall quality~\cite{Knoche_2020_CVPRW}.

In terms of architecture, the model is defined as a conditional generative adversarial network (GAN) ~\cite{goodfellow2014generative} with the pose and image as input. The input can be represented as 2D pose~\cite{Ma_2017_NIPS}, semantic segmentation~\cite{Xu_2023_TMM}, or even UV- maps~\cite{guler2018densepose}. UNet~\cite{Xiaomeng_2018_TMI} is largely used with residual blocks~\cite{Kaiming_2016_CVPR} as main convolutional modules. Since the input and target poses may differ in terms of scale, orientation, and spatial position, various solutions have been proposed to account for this. Deformable GAN~\cite{Siarohin2017DeformableGF} does this by using a mask extracted from the target pose and applying it to the UNet skip connection in order to force the feature to focus on the spatial positioning of the pose, thus alleviating the problem of pixel misalignment. We can categorise the methods to tackle the input-to-target pose into warping and progressive methods. Warping methods~\cite{han2017viton,Siarohin_2019_CVPR} use warping fields to transform (or warp) the feature from the input using the target pose as guidance. Progressive methods~\cite{Zhu_20202_TPAMI,tang2022multi} attempt to transfer spatial regions (or patches) from the input pose to the target pose.


\begin{figure*}
    \centering
    \includegraphics[width=1.\textwidth]{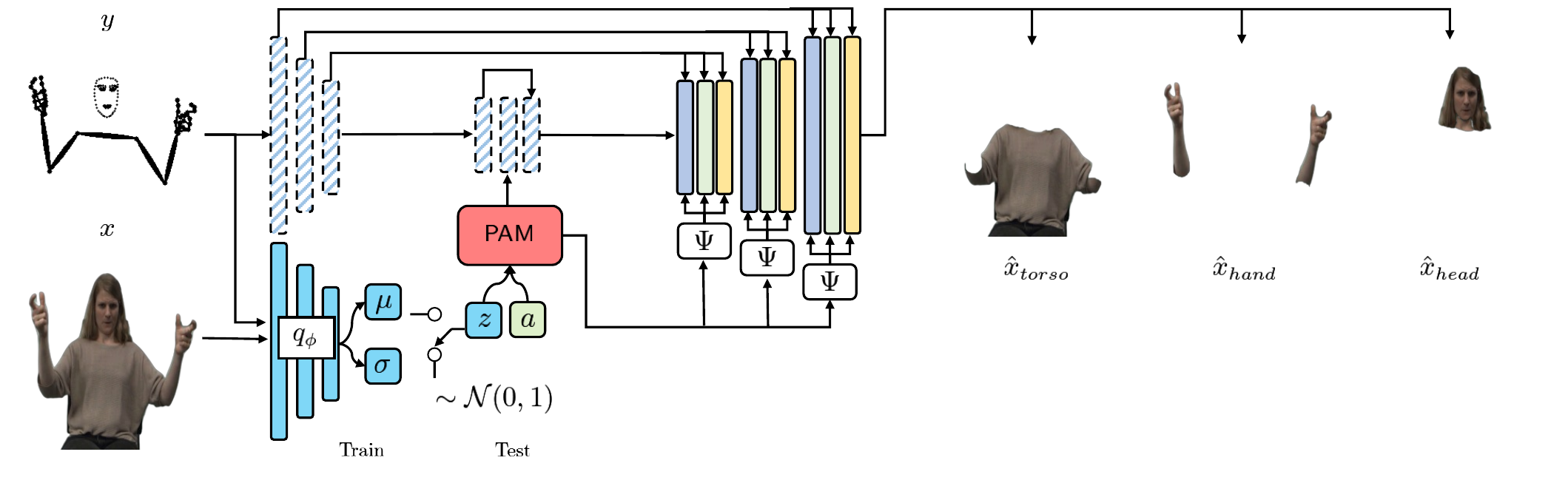}
    \caption{\textbf{PENet.} The network is presented as a conditional VAE-GAN, where the variational parts learn the distribution of visual feature from signers of different attributes (skin tone, ethnicity, gender) through variational inference. The attribute $a$ is presented as a feature vector extracted from a pre-trained CLIP model. The latent code $z$ and $a$ are combined through a MHA module. The pose $y$ is processed through a UNet encoder-decoder network to retain the spacial information of the keypoints, the visual feature $z_a$ guides the synthesis of the person through a mapping $\Psi$ (Eq.~\ref{psi_z}).}
    \label{fig:PE_VAE_main_pipeline}
\end{figure*}

\section{Method}  \label{sec:met} This section first provides a problem formulation for diversity-aware Sign Language Production. Then, we relate the use of Variational Auto-Encoder~\cite{Diederik_2019_Arxiv} and Vision Transformers~\cite{dosovitskiy2021an} to our proposed framework. In particular, we introduce Pose-Encoding Variational Inference (PE-VAE), which explicitly uses the pose and attributes as conditioning in the lower bound computation. The proposed PE-VAE enables us to learn better pose-agnostic features. Finally, we introduce our Pose-Encoding Network (PENet), a UNet encoder-decoder model with multi-head decoders representing each of the head, hand, and torso. PENet takes an input pose and generates an anonymized person based on the conditional attribute.

\subsection{Problem definition} Given a sequence consisting of $T$ RGB frames \mbox{$\mathcal{X} = \{ x_t \in \mathbb{R}^{W \times H \times 3} \}_{t=1}^{T}$} of a deaf signer, and their corresponding $2$D pose $\mathcal{Y} = \{ y_t \in \mathbb{R}^{K\times 2} \}_{t=1}^{T}$ where $K$ is the number of joints, $W$ (resp. $H$) is the image width (resp. height).
We formalise the Diversity-Aware Sign Language Production framework with the help of a generator \mbox{$\mathbf{G}: \mathbb{R}^{K \times 2} \times \mathbb{R}^{d} \to \mathbb{R}^{W \times H \times 3} $}, which takes a pose $y \in \mathcal{Y}$ together with an attribute $a \in \mathbb{R}^d$ of feature dimension $d$, and maps this to the pixel space $x \in \mathcal{X}$, such that: $x \approx \mathbf{G}(x|y, a)$. We do not consider a temporal consistency term since the pose sequence $\mathcal{Y}$ contains the motion information.

\subsection{Preliminaries}
\noindent\textbf{Conditional Variational Auto-Encoder.} A conditional variational auto-encoder (cVAE) ~\cite{Diederik_2019_Arxiv} defines a two-stage process: generative and inference. The generative, or encoding, process learns a latent code $z$ from an input $x$ so that the prior distribution is $p_\theta(z|x)$. The latent code $z$ is used to infer the input $x$ during inference or decoding. Because of intractability, we approximate the true posterior with the mapping $q_\phi(z|x)$ also called the variational posterior.
By variational inference~\cite{Diederik_2019_Arxiv}, the evidence lower bound of the log-likelihood term is:
\begin{equation}
\begin{aligned}
\label{conditional_likelihood}
    \log p_\theta(x) \geq  \mathbb{E}_{q_\phi(z|x)}\log p_\theta(x|z) - \\ \beta D_{KL}(q_\phi(z|x)\|p(z)).
\end{aligned}
\end{equation} The prior $p(z)$ is given as a Gaussian $\mathcal{N}(0, 1)$ and $p_\theta(x|z)$ is defined as the reconstruction term, which is controlled by pixel-wise supervision.
The $\beta$ term controls the estimation of the posterior with the chosen prior~\cite{higgins2017betavae}.

\textbf{Vision Transformer.} The Vision Transformer (ViT) presented in~\cite{dosovitskiy2021an} partitions an image $x$ into $N = \lfloor \frac{H}{h} \rfloor \times \lfloor \frac{W}{w} \rfloor$ non-overlapping patches $x_1, \dots, x_N \in \mathbb{R}^{h\times w \times c}$, where $w, h$ are the width and height of the patch respectively. The patches are used as tokens by a linear projection for each $x_i$ as $\mathbf{z}_i \in \mathbb{R}^d$ using a linear operator $\mathbf{E}$, as $\mathbf{z}_i = \mathbf{E}x_i$.

The tokens are then concatenated into a sequence, together with an additional learnable class token $\mathbf{z}_{cls} \in \mathbb{R}^{d}$~\cite{devlin_naacl_2019}. A positional embedding $\mathbf{p} \in \mathbb{R}^{(N + 1) \times d}$, is also added to the sequence. The tokenization process is thus given as follows:
\begin{equation}
    \mathbf{z}^{0} = [\mathbf{z}_{cls}, \mathbf{E}x_1, \mathbf{E}x_2, \ldots, \mathbf{E}x_N] + \mathbf{p}.
\end{equation}

The token sequence $\mathbf{z}$ is applied to a transformer encoder with $L$ layers.
Each layer, $\ell$, is defined as,
\begin{align}
 \mathbf{y}^{\ell} &= \text{Att}\left(\text{LN}\left(\mathbf{z}^{\ell - 1}\right)\right) + \mathbf{z}^{\ell - 1}, \label{eq:transformer_layer_msa} \\
 \mathbf{z}^{\ell} &= \text{MLP}\left(\text{LN}\left(\mathbf{y}^{\ell}\right) \right) + \mathbf{y}^{\ell}.
 \label{eq:transformer_layer_mlp}
\end{align} Where Att is the attention operation using the Query (Q), Key (K), and Value (V) as defined in~\cite{NIPS2017_3f5ee243}, respectively, LN represents layer norm~\cite{ba2016layer} and MLP is a feed-forward neural network.

\subsection{Pose-Encoding Variational Inference} \label{sec:PEVAE}
In the context of diversity-aware SLP, we encourage the generative process to learn the latent code $z$ from the image $x$, the pose $y$, and the attribute condition $a$. We modify the prior distribution as $p_\theta(z|x, y, a)$. Therefore, the pose-conditioned evidence lower bound of the log-likelihood term is updated such that:
\begin{equation} 
\begin{aligned}
\label{conditional_likelihood_updated}
    \log p_\theta(x| y, a) \geq  \mathbb{E}_{q_\phi(z|x, y, a)}\log p_\theta(x|z, y, a) - \\ \beta D_{KL}(q_\phi(z|x, y, a)\|p(z)).
\end{aligned}
\end{equation}

To implement the conditional posterior $q_\phi$, defined in Eq.~\ref{conditional_likelihood_updated} we use separate encoders for both the pose $y$ and the image $x$. The feature representation of each input is then combined using a multi-head attention module as defined in Eq.~$(2)$-$(4)$ such that: 
\begin{equation} 
\begin{aligned}
\label{eq:za_feat}
    \mathbf{f}_{xy} = \text{MHA}(\mathcal{E}_x(x) \oplus \mathcal{E}_y(y))), \\ \
    \mu = \text{MLP}(\mathbf{f}_{xy}), \sigma = \text{MLP}(\mathbf{f}_{xy}).
\end{aligned}
\end{equation}
Where $\mathcal{E}_x$ (resp. $\mathcal{E}_y$) is the image (resp. pose) encoder.
Compared to the previous state-of-the-art~\cite{Saunders_2021_FG}, where the conditioning $y$ is omitted during posterior training, we argue that this simplification is suboptimal. Indeed, the latent code $z$ encapsulates the visual features of the signer, and thus, such a representation must be independent of the pose. Therefore, explicitly including the pose $y$ in the conditioning helps to obtain better visual features. Finally, we obtain the latent code by reparameterization as $z = \mu + \sigma \epsilon$, where $\epsilon \sim \mathcal{N}(0, 1)$.

We assume that $\mathcal{N(\mu, \sigma)}$ is an uncorrelated multivariate Gaussian random variable of dimension $M$. Therefore, the VAE loss is given as:
\begin{equation} 
\begin{aligned}
    L_{\textbf{VAE}} = \frac{1}{2} \sum_{i=1}^{M}(1+\log(\sigma_i^2)-   \sigma_i^2-\mu_i^2).
\end{aligned}
\end{equation}

\subsection{Pose Encoding Model (PENet)} 
\begin{figure}
    \centering
    \includegraphics[width=.43\textwidth]{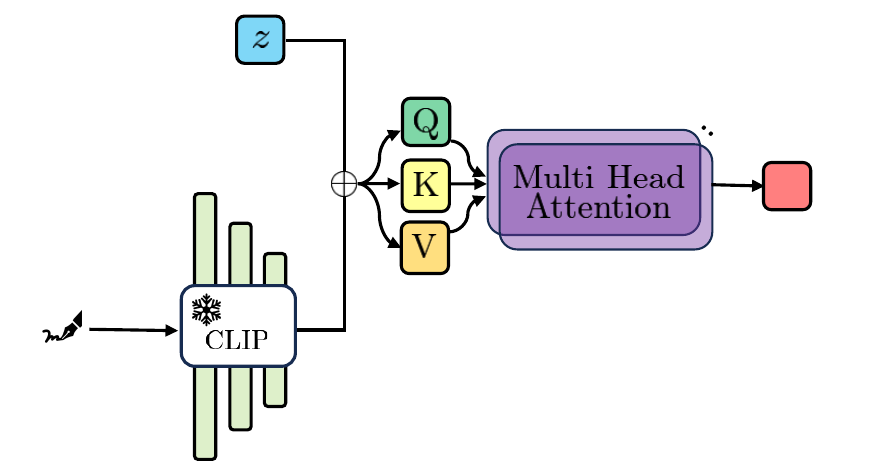}
    \caption{\textbf{Pose Aggregation Module}. Using a text prompt of an attribute (\textit{e.g.,} gender), we extract a feature $a \in \mathbb{R}^{512}$ from a pre-trained CLIP model~\cite{Radford_2021_arxiv}. This feature is then concatenated with the latent code $z$ and fed into a multi-layer attention module.}
    \label{fig:attrib_fusion}
\end{figure}

\textbf{Model description.}
We build our generator $\mathbf{G}$ using a UNet architecture~\cite{Xiaomeng_2018_TMI}. The reason is that such a class of models retains spatial information from the encoder through skip connections, which is helpful since the network uses the pose $y$ to synthesize an image from the latent code. The latent feature $z$ is combined with the attribute feature $a = \text{\footnotesize \sffamily{CLIP}(attribute)}\footnote{We use a predefined text prompt for each attribute} \in \mathbb{R}^{512}$, which is obtained by a pre-trained CLIP model (Fig.~\ref{fig:attrib_fusion}), so that: \mbox{$\mathbf{z}_a = \text{MHA}(z, a)$}.

The skip connection of a UNet model with $L$ layers is defined as the concatenation between the feature encoder $\mathbf{f}_{\mathcal{E}}^i$ on layer $i$ and the symmetric decoder feature $\mathbf{f}_{\mathcal{D}}^{L-i}$ on the symmetric decoder layer $L-i$.

In the pose-guided human image synthesis problem, the input to the generator is an image with the target pose; in such a case, the encoder contains visual features. In diversity-aware SLP, the encoder contains only coarse keypoint features, making it difficult for the network to learn meaningful visual appearance features. A simple solution is to combine the latent code $\mathbf{z}_a$ with the last encoding feature in the bottleneck. Such a way might not be efficient since the bottleneck layer loses spatial information. Therefore, we propose to inject $\mathbf{z}_a$ into all decoding layers using a mapping $\Psi$. The skip connection is then defined as follows:

\begin{equation} 
\begin{aligned}
\label{psi_z}
 \mathbf{f}_{\text{skip}}^{L-i} = \Psi( \mathbf{f}_{\mathcal{E}}^i \oplus \mathbf{f}_{\mathcal{D}}^{L-i}, \mathbf{z}_a).
\end{aligned}
\end{equation}

To capture nonmanual features, we opt for a separate head decoder. To retain the consistency of the different body parts, we share the weights across the encoder and the bottleneck layers. The predicted image is therefore obtained as follows:

\begin{equation} 
\begin{aligned}
\label{UNET_skip_connection}
 \hat{x} = \hat{x}_{hand} . \mathbf{m}_{hand} + \hat{x}_{head} . \mathbf{m}_{head} + \hat{x}_{torso} . \mathbf{m}_{torso},
\end{aligned}
\end{equation}
where $\mathbf{m}_{hand}, \mathbf{m}_{head}, $ and $\mathbf{m}_{torso}$ are the hand, head, and torso masks, respectively. \textcolor{green}{}

\textbf{Losses.}  We use GAN optimisation~\cite{goodfellow2014generative} to train our model. For the generator, we use Perceptual loss~\cite{Johnson_2016_ECCV} to account for low and high frequencies. It uses a VGG$_{19}$~\cite{Simonyan15} model and maps the input to a set of $L^{V}$ higher feature spaces and penalizes the prediction $\hat{x}$ in these spaces. The loss is given as:

\begin{equation}
    L_{\textbf{perc}} = \sum_{l=1}^{L^{V}} \norm{\text{VGG}^l_{19}(x) - \text{VGG}^l_{19}(\hat{x})}_2.
\end{equation} 

To further improve for higher frequency details, we use an edge loss that utilizes a combination of Sobel~\cite{Kanopoulos_1988_sobel}, Laplacian~\cite{Burt_1983_lap} filters, and Canny~\cite{Canny_1986_PAMI} edge detection. The edge loss is then:
\begin{equation}
    L_{\textbf{edge}} = \text{Canny}(x, \hat{x}) + \text{Laplacian}(x, \hat{x}) + \text{Sobel}(x, \hat{x}).
\end{equation} 

In addition, we use a multi-scale discriminator loss~\cite{wang2018pix2pixHD} with $L^{\mathbf{D}}$ layers, which forces the generator $\mathbf{G}$ to output realistic predictions through a min-max optimization. We also use a conditional discriminator to allow the network to focus on each body part separately. Instead of using a separate discriminator for each body part, we merge them into a single input. The loss is then calculated as follows:

\begin{equation}
    L_{\textbf{feat}} = \sum_{l=1}^{L^{\mathbf{D}}} \norm{ \text{\footnotesize \sffamily{tr}}^{l} - \text{\footnotesize \sffamily{pr}}^{l} },
\end{equation} 

where $\text{\footnotesize \sffamily{tr}} = \log \mathbf{D}(y, x_{hand} \oplus x_{head} \oplus x_{torso})$ is the discriminator output from the true sample. The discriminator output from the synthesis of the three parts is  \mbox{$\text{\footnotesize \sffamily{pr}} = \log(1 - \mathbf{D}(y, \hat{x}_{hand} \oplus \hat{x}_{head} \oplus \hat{x}_{torso} ))$}.

We use an attribute loss; the idea is to force the weights of $\mathbf{G}$ to learn discriminative features toward the conditional attribute $a$. We therefore use a pre-trained ResNet-$18$~\cite{Kaiming_2016_CVPR} where we freeze its weights and add a classification layer. We use cross-entropy to implement the loss $L_{\textbf{attrib}}$.

In summary, the total loss to train the generator is given as: \mbox{$L = L_{\textbf{perc}} + L_{\textbf{feat}} + \lambda_{\textbf{edge}} L_{\textbf{edge}} + \lambda_{\textbf{attrib}} L_{\textbf{attrib}} + \beta L_{\textbf{VAE}}$}, where $\lambda_{\textbf{edge}}$, $\lambda_{\textbf{attrib}}, \beta$ are weighting terms.

\begin{table*}[t!]
		\caption{State-of-the-art comparison with ablation study of our proposed PENet. KEY -- Colors: \textcolor{p_tab_best}{\textbf{best}}, \textcolor{p_tab_sbest}{second-best}.  * We use the original implementation as proposed by the authors.}
    \centering
    \scriptsize
    \label{tab:res}
    \resizebox{\textwidth}{!}{\begin{tabular}{l|ccc|ccc|ccc}
        \hline
        \multirow{2}{*}{\textbf{Model}} & \multicolumn{3}{c}{$\uparrow$ \textbf{SSIM}}  & \multicolumn{3}{c}{$\uparrow$ \textbf{PSNR}}  & \multicolumn{3}{c}{$\downarrow$ \textbf{FID}} \\ 
        & Head & Hand & Torso & Head & Hand & Torso & Head & Hand & Torso \\ \hline
        Libras~\cite{Silveira_2022_SIBGRAPI} & $.957 \pm .006$ & $.936 \pm .020$ & $.873 \pm .014$ & $23.32 \pm 1.00$ & $21.07 \pm 1.58$ & $16.42 \pm 1.28$ & $334.12 \pm 25.42$ & $306.55 \pm 24.42$ & $264.92 \pm 23.56$ \\
        Anonysign~\cite{Saunders_2021_FG}* & $.976 \pm .007$ & $.959 \pm .015$ & $.944 \pm .027$ & $30.58 \pm 2.00$ & $26.61 \pm 2.41$ & $27.79 \pm 3.57$ & $68.11 \pm 23.40$ & $127.51 \pm 20.82$ & $147.80 \pm 25.68$ \\ \hline
        \multirow{11}{*}{\rotatebox[origin=c]{90}{PENet}} \phantom{a} w/o $L_{\textbf{edge}}$  & $.980 \pm .005$ & $.971 \pm .012$ & $.949 \pm .026$ & $32.55 \pm 2.09$ & $29.26 \pm 2.31$ & $29.61 \pm 3.77$ & $55.09 \pm 18.71$ & $88.65 \pm 15.64$ & $133.70 \pm 23.57$ \\
         \phantom{aaa} $L_{\textbf{edge}}$ & $\secbest{.984 \pm .005}$ & $.970 \pm .012$ & $.947 \pm .026$ & $33.69 \pm 2.39$ & $29.12 \pm 2.22$ & $29.33 \pm 3.53$ & $46.90 \pm 19.07$ & $83.28 \pm 16.75$ & $128.06 \pm 23.21$ \\ 
         \phantom{aaa} $\mathbf{f}_{xy}$ (\text{\tiny \sffamily{Conv}}) & $.983 \pm .005$ & $.968 \pm .014$ & $.948 \pm .028$ & $33.35 \pm 2.55$ & $28.81 \pm 2.54$ & $29.53 \pm 3.84$ & $55.34 \pm 19.57$ & $95.45 \pm 16.59$ & $138.05 \pm 27.91$ \\ \cdashline{2-9}

         \phantom{aaa} PE (concate) & $\secbest{.984 \pm .005}$ & $.970 \pm .014$ & $.945 \pm .027$ & $33.72 \pm 2.88$ & $28.98 \pm 2.73$ & $28.06 \pm 3.60$ & $51.96 \pm 21.46$ & $86.52 \pm 19.23$ & $134.08 \pm 28.54$ \\
         \phantom{aaa} PE (shared) & $\secbest{.984 \pm .005}$ & $.972 \pm .013$ & $.950 \pm .028$ & $34.04 \pm 2.74$ & $29.77 \pm 2.50$ & $29.99 \pm 4.11$ & $59.76 \pm 27.21$ & $83.70 \pm 15.67$ & $129.01 \pm 23.99$ \\
         \phantom{aaa} PE (separate) & $\best{.985 \pm .005}$ & $.972 \pm .012$ & $\secbest{.951 \pm .003}$ & \secbest{$33.98 \pm 2.65$} & $29.67 \pm 2.46$ & $\secbest{30.33 \pm 3.84}$ & $\best{48.37 \pm 19.41}$ & $\secbest{77.10 \pm 17.67}$ & $\best{125.23 \pm 23.84}$ \\ \cdashline{2-9}

         \phantom{aaa} w/o $\mathbf{f}_{\text{skip}}$ & $.950 \pm .007$ & $.912 \pm .028$ & $.838 \pm .025$ & $24.20 \pm 1.27$ & $20.55 \pm 1.57$ & $13.77 \pm 1.80$ & $224.28 \pm 30.42$ & $306.28 \pm 30.25$ & $303.46 \pm 34.77$ \\
         \phantom{aaa} w/o $\mathbf{f}_{\text{skip}} \& \Psi$ & $.947 \pm .007$ & $.916 \pm .031$ & $.810 \pm .020$ & $23.22 \pm 1.03$ & $20.33 \pm 1.40$ & $12.16 \pm 1.96$ & $266.48 \pm 28.84$ & $339.12 \pm 31.98$ & $307.93 \pm 33.22$ \\
         \phantom{aaa} $\Psi$(\text{\tiny \sffamily{Conv}}) & $\best{.985 \pm .005}$ & $\secbest{.973 \pm .012}$ & $.950 \pm .029$ & $\best{34.21 \pm 2.63}$ & $\secbest{30.08 \pm 2.24}$ & $29.90 \pm 4.10$ & $54.56 \pm 23.65$ & $82.58 \pm 17.20$ & $132.21 \pm 26.00$ \\ \cdashline{2-9}
         \phantom{aaa} w/o ($\mathbf{m}_{hand}$) & $.982 \pm .006$ & $.964 \pm .012$ & $.947 \pm .026$ & $32.18 \pm 3.01$ & $27.63 \pm 1.76$ & $26.70 \pm 3.13$ & $117.88 \pm 37.02$ & $147.30 \pm 20.89$ & $180.86 \pm 34.64$ \\
         \phantom{aaa} full model & $\secbest{.984 \pm .005}$ & $\best{.983 \pm .006}$ & $\best{.956 \pm .027}$ & $33.80 \pm 2.66$ & $\best{33.38 \pm 1.93}$ & $\best{31.44 \pm 4.51}$ & $\secbest{49.92 \pm 15.96}$ & $\best{72.07 \pm 30.22}$ & $\secbest{125.29 \pm 25.37}$ \\
        \hline
    \end{tabular}}
\end{table*}

\section{Experiments}
\label{sec:exp}
In this section, we evaluate the performance of PENet. We outline the experimental setup and then quantitatively and qualitatively evaluate our model against solid baselines.

\subsection{Experimental setup}
\textbf{Dataset:}  Due to the difficulty in collecting datasets that are suitable to our need, we choose to use the SMILE~\cite{SMILE_2018_LREC} sign language dataset  as it has a wide range of glosses. Furthermore, we collect additional videos from YouTube to add signers for the ethnicity and skin color attributes. We further process the data containing only atomic glosses (\textit{i.e.} discarding idle frames). We use a weighted data sampler because of the highly imbalanced data distribution. As a data augmentation technique, we add a random rotation of up to 15 pixels, shift, scale, and random horizontal flipping.

\textbf{Implementation Details:} To train PENet, we set \mbox{$\lambda_{\textbf{edge}}=0.01$}, \mbox{$\lambda_{\textbf{attrib}}=0.001$}, $\beta=0.001$ and we keep this value fixed throughout the paper unless otherwise specified.

Since no ground truth is available for the attributes (gender, skin tone, and ethnicity), we test the performance of our model in an unconditional setting. Therefore, we drop the attribute and use supervised pixel-to-pixel metrics. In particular, we use Structural Similarity (SSIM), Peak Signal-to-Noise-Ratio (PSNR)~\cite{Wang_2004_TIP} as per-pixel measurement, and Fr\'echet Inception Distance (FID)~\cite{Martin_2017_NIPS} to assess the diversity of our generative model. To measure the performance of the non-manual features, we use the masked version of each metric for the head, hand, and torso. We also add a pose estimation metric where for each input pose, we sample $5$ random samples and run the estimation. We use $\texttt{Mediapipe}$~\cite{Camillo_2019_arxiv} as a pose estimation model. Furthermore, we add the success rate (denoted by Hit$@$) when the pose model successfully estimated the keypoint for the synthesis and the ground truth.

\subsection{Ablation study}
\textbf{Pose representation.} The heatmap representation is widely adopted in the literature~\cite{Ma_2017_NIPS,Saunders_ArXiv_2020}. Subsequently, it is the default representation for Sign Language Given a pose $y = \{y_1, \dots, y_K\}$ 
of $K$ joints,  the heatmap $\mathcal{H}$ is defined as: \begin{equation} \label{eq:H}
\mathcal{H}_i(y) = \exp \Big(- \frac{||p - y_i||^2_2}{2\tau^2} \Big),
\end{equation} where $p$ is the pixel position for an RGB image of size $\mathbb{R}^{3 \times W \times H}$ and $\tau=6$.

Instead, we propose using RGB-like skeleton images and empirically found that it provides better image quality while considerably reducing the training time by about a half. 

\begin{figure}[h!]
    \centering
    \includegraphics[width=.45\textwidth]{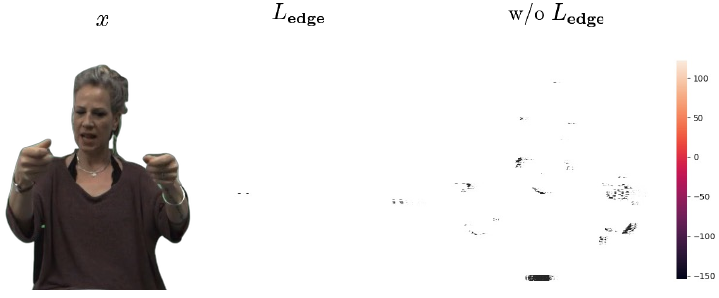}
    \caption{\textbf{Edge loss}. Effect of the edge loss on the synthesised frames. We show the heatmaps of the same frame with and without using the edge loss $L_{\textbf{edge}}$. Notice the errors when using $L_{\textbf{edge}}$ come mainly from the hand of the clothes \textit{i.e.}, not in the boundaries of the body.}
    \label{fig:edge_loss_res}
\end{figure}
\textbf{Edge loss.} Adding the edge loss $L_{\textbf{edge}}$ contributes to the high-frequency detail and produces crisper results. From Tab.~\ref{tab:res}, we can see that adding the proposed edge loss provides better qualitative results, which are manifested by a lower FID scores against the baseline model. In Fig.~\ref{fig:edge_loss_res}, we can see that the edge loss has no errors on the body boundaries. Also, we note that the face details are much more visible when using $L_{\textbf{edge}}$.

\textbf{Feature representation.} To represent $\mathbf{f}_{xy}$, we compare the proposed multi-layer attention module as a feature aggregator in Eq.~\ref{eq:za_feat} against convolution. The convolution module noted as \mbox{$\mathbf{f}_{xy}$ (\text{\tiny \sffamily{Conv}})} is defined as a convolution layer with a kernel of size $3$, stride of $1$, and padding of $1$ followed by a LeakyReLU with a negative slope of $0.2$. Results from Tab.~\ref{tab:res} show that the proposed MHA (second row in PENet) produces better results than the convolution module. 
Furthermore, we also run a similar experiment for the attribute aggregation Eq.~\ref{eq:za_feat}, and we replace the MHA with a Linear layer to match the dimension. We note similar behaviors as above.

One reason to explain this is the change of each representation. When using CNNs, the kernels are fixed and applied to the input directly, whereas when using MHA, we take into account the input through Query/Key/Value matrices. Hence, since the latent feature $z$ is randomly sampled through a Gaussian at test time, it is more convenient to attend (or weight) the importance of the learned matrices.

\begin{figure}[h!]
    \centering
    \includegraphics[width=.47\textwidth]{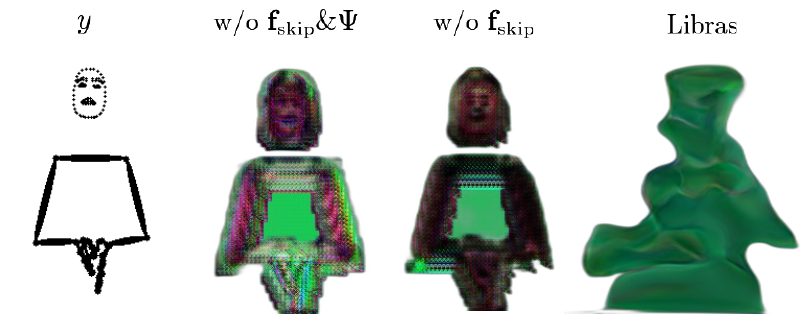}
    \caption{Example showing the effect of skip connections and injecting the appearance feature $\mathbf{z}_a$ using $\Psi$ in the decoder.}
    \label{fig:noskip_bad}
\end{figure}

\textbf{Unet representation.} The UNet can retain spatial information~\cite{Xiaomeng_2018_TMI}. In the context of our problem, the network must use skip connections to focus only on painting the semantic body regions. To test this hypothesis, we ran our baseline by removing skip connections; we found that the network could not train properly, and the images consisted of only flickering patches around the poses (Fig.~\ref{fig:noskip_bad}). Similar behavior was noticed when training Libras~\cite{Silveira_2022_SIBGRAPI}.

\textbf{Feature aggregator $\Psi$.} We use a StyleGAN~\cite{Karras_2019_CVPR} feature aggregation to implement $\Psi$. Additionally, we remove the module in the decoding step noted as $\Psi$(\text{\tiny \sffamily{Conv}}). From Tab.~\ref{tab:res}, we see that the convolution baseline produces better scores for the head with $.985$ vs. $.984$ SSIM and $34.21$ vs. $33.80$ PSNR. However, our $\Psi$ provides better scores on average, which are translated into the FID scores and better synthesis in general. 

\textbf{Pose-encoding VAE.} As explained in Sec.~\ref{sec:PEVAE}, we explicitly inject the pose information $y$ into the ELBO calculation. Since there is no ground-truth conditional latent distribution, we relax it to a normal Gaussian distribution $\mathcal{N}(0, 1)$. However, we retain the conditioning on the posterior estimation of $q_\phi$. In the following, we empirically verify that retaining the conditioning helps to learn better agnostic visual features $\mathbf{z}_a$.

We propose a late fusion scheme to combine the pose $y$ and the visual feature from the image $x$. We propose the following ablation to show the effectiveness of such a scheme.

\textit{Early-fusion:} The pose and the image are concatenated together $y \oplus x$ before being fed to a single encoder that implements $q_\phi$.

\textit{Shared encoder:} The weights of the pose encoder are shared with the pose encoder of the generator $\mathbf{G}$. The image $x$ has its encoder, and then the feature representation of both inputs is first concatenated and then fed to a multi-layer attention module with $2$ layers.

\begin{figure}[h!]
    \centering
    \includegraphics[width=.45\textwidth]{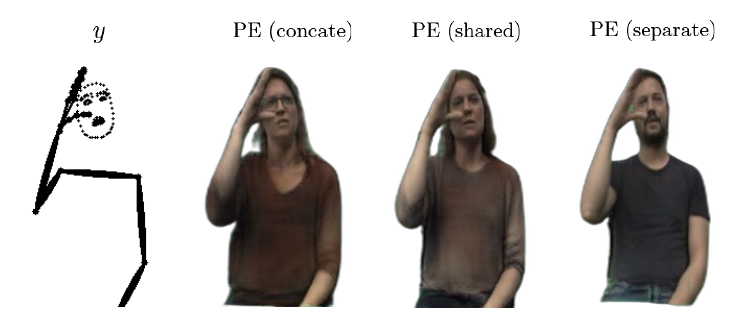}
    \caption{Results showing various pose encoding VAE schemes as presented in Sec.~\ref{sec:PEVAE}.}
    \label{fig:PE_VAE}
\end{figure}

\textit{Late-fusion:} We follow the same architecture as the shared encoder, except that we do not share the weights of the pose encoder.
We note that the three schemes produce similar qualitative results overall. With some differences, as can be seen in Fig.~\ref{fig:PE_VAE}. In particular, we found more visible artifacts in the early fusion. This is because $x$ and $y$ are mixed within the same encoder. We see fewer artifacts when we use separate encoders, as with PE(shared) and PE(separate). Using a dedicated pose encoder for the generator $\mathbf{G}$ and $q_\phi$ provides better qualitative results.

\begin{table}[t!]
		\caption{Pose estimation results. For each input pose $y$ we sample $z \sim \mathcal{N}(0,1), 5$ times and then run the pose estimator.}
    \centering
    \scriptsize
    \label{tab:res_kps}
    \begin{tabular}{l|ccr}
        \hline
         & AnonySign  & PENet & Gain\phantom{aa} \\  \hline
        
        \multirow{4}{*}{$L_2$} Head & $2.21 \pm 3.86$ & $\textbf{1.13} \pm \textbf{3.43}$ & $\textbf{1.08}\downarrow$\\
        \phantom{aaa} R-Hand & $21.61 \pm 35.72$ & $\textbf{3.72} \pm \textbf{13.32}$ & $\textbf{17.89}\downarrow$ \\ 
        \phantom{aaa} L-Hand & $12.29 \pm 22.21$ & $\textbf{4.89} \pm \textbf{16.49}$ & $\textbf{7.40}\downarrow$ \\ 
        \phantom{aaa} Clothes & $4.45 \pm 6.34$ & $\textbf{2.81} \pm \textbf{4.66}$ & $\textbf{1.64}\downarrow$ \\ \hline
        \multirow{4}{*}{\rotatebox[origin=c]{90}{Hit$@$}} Head & $90.87\%$ & $\textbf{91.29\%}$ & $\textbf{0.42}\uparrow$ \\
        \phantom{aa} R-Hand & $8.28\%$ & $\textbf{38.25\%}$ & $\textbf{29.97}\uparrow$  \\ 
        \phantom{aa} L-Hand & $14.44\%$ & $\textbf{22.03\%}$ & $\textbf{7.59}\uparrow$  \\ 
       \phantom{aa} Clothes & $\textbf{99.58\%}$ & $99.35\%$ & $0.23\downarrow$  \\ \hline
    \end{tabular}
\end{table}

\subsection{State-of-the-art comparison} 
\begin{figure*}[h!]
    \centering
    \includegraphics[width=.8\textwidth]{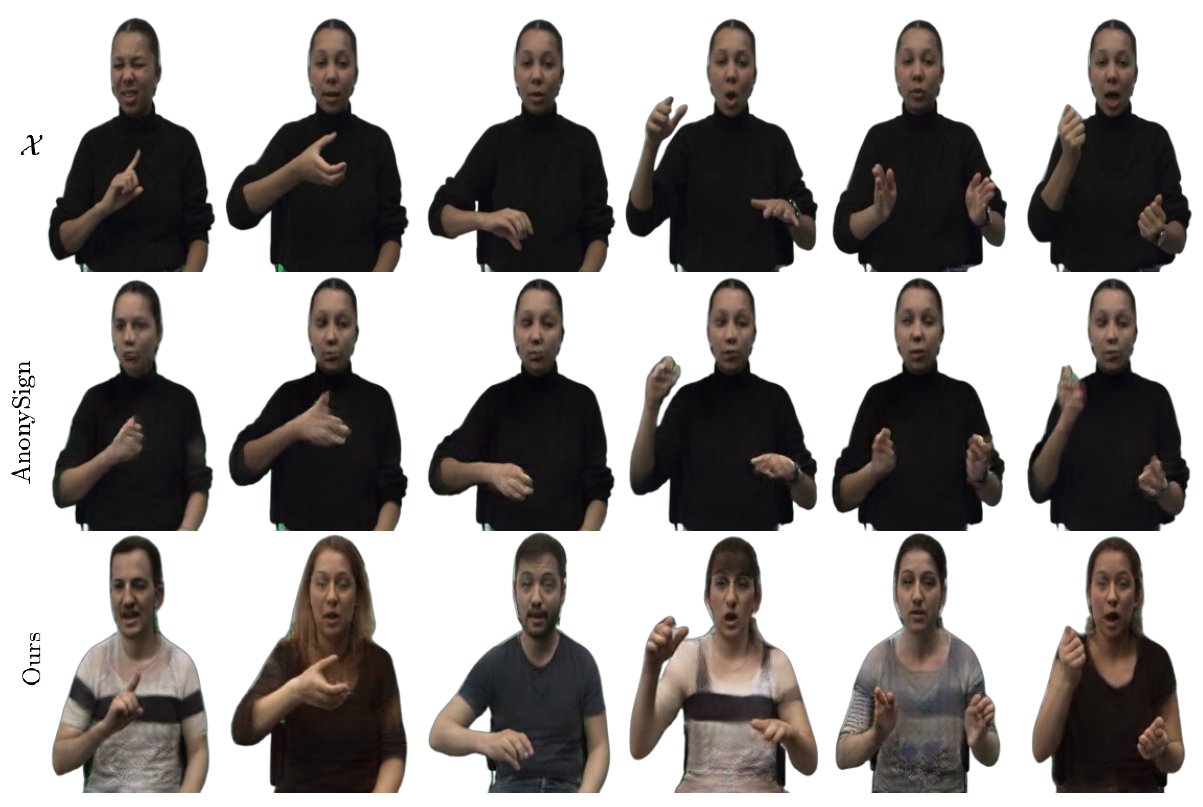}
    \caption{Method comparison of our proposed PENet against AnonySign~\cite{Saunders_2021_FG} using the pose sequence as conditioning.}
    \label{fig:soa_cmp}
\end{figure*}

Our proposed PENet addresses the shortcomings of the two state-of-the-art methods we compare against. The stability of the synthesis as in Libras~\cite{Silveira_2022_SIBGRAPI} and the diversity of samples from which AnonySign~\cite{Saunders_2021_FG} suffers (Fig.~\ref{fig:soa_cmp}).

We address the stability via UNet skip connections as shown in Fig.~\ref{fig:noskip_bad}. As for the diversity of the samples, we found that the style loss used in~\cite{Saunders_2021_FG} pushes the network to reduce diversity and focus more on the reconstruction loss. Therefore, we decided not to use such a loss in our model.

In Fig.~\ref{fig:soa_cmp}, we compare the samples generated by our model with AnonySign. As can be seen, our model has better image diversity and better non-manual features. In particular, our PENet shows distinct facial expressions and accurate hands thanks to our pose-encoding VAE approach.

From Tab.~\ref{tab:res}, we note that the multi-branch approach gives the best results overall. However, we found that adding the hand segmentation helps the network to produce more accurate results. The hand segmentation acts as an explicit mask for each decoder.

\begin{figure}[h!]
    \centering
    \includegraphics[width=.35\textwidth]{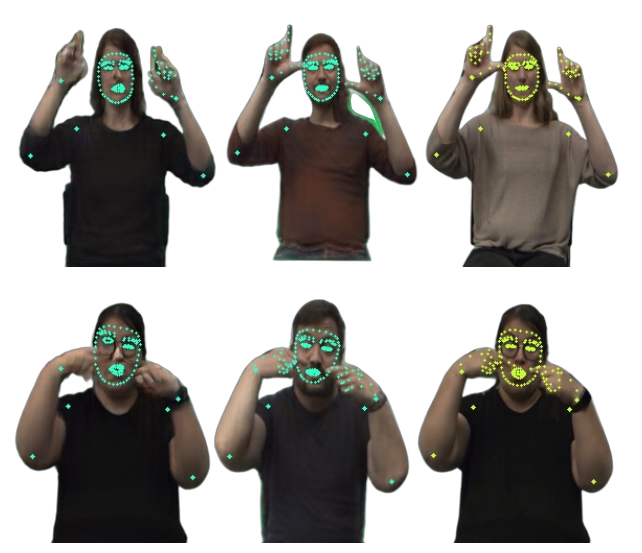}
    \caption{Qualitative comparison of the pose estimation on the synthesised images of signers. The keypoint estimation is highlighted in colors on top of the image. KEY -- left: Anonysign; middle: PENet (ours); right: ground-truth.}
    \label{fig:kps_cmp}
\end{figure}

The results from Tab.~\ref{tab:res_kps} show that our proposed PENet outperforms AnonySign in skeleton estimation, especially in hand estimation, where our model has $3.72$ (resp. $4.89$) pixel error for the right (resp. left) hand. AnonySign, on the other hand, has a pixel error of $21.61$ (resp. $12.29$) pixel error. This is a significant improvement and shows that our model is robust, and we can rely on it for other tasks, such as data augmentation. We also note that our model has a pixel error of $1.13$ for the head, which, as mentioned earlier, is suitable for capturing non-manual features. We also found that the Hit$@$ rate for the hands is lower. This is because the bounding boxes around the hand are small (less than $40 \times 40$), and the pose estimator sometimes cannot estimate the hands even for the ground-truth images. Fig.~\ref{fig:kps_cmp} shows the typical behavior of the pose estimation.

\subsection{Qualitative analysis} As highlighted earlier, we have no ground-truth samples for the attribute-based synthesis. The main difference with the unconditional synthesis is the additional attribute input. Note that in this case, we replace $y$ in Eq.~\ref{eq:za_feat} with $y \oplus a\uparrow$ where $a\uparrow$ is implemented as a $4$-layer MLP followed by a $3$-layer transpose convolution in order to match with the input image dimension of shape $\mathbb{R}^{256 \times 256 \times 3}$. Doing this allows for better modeling of the conditional attribute.

\begin{figure*}[h!]
    \centering
    \includegraphics[width=.7\textwidth]{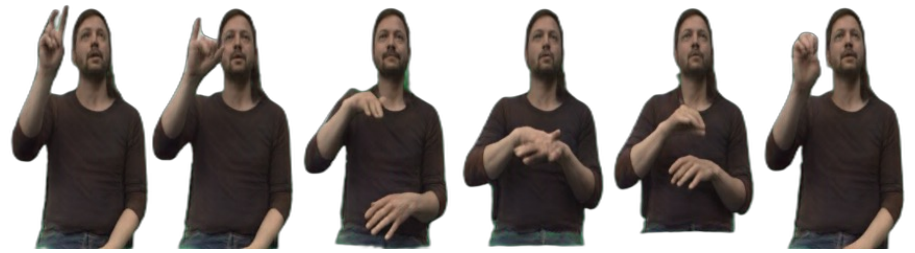}
    \caption{\textbf{Synthesis consistency}. By fixing the latent attribute feature $\mathbf{z}_a$, our model can synthesize consistent identities with the skeleton conditioning. This example shows a synthesis from female $\to$ male.   }
    \label{fig:male}
\end{figure*}
\textbf{Consistency.} We fix the latent feature $\mathbf{z}_a$ by randomly sampling from a normal distribution for the input sequence $\mathcal{Y} = \{y_1, \dots, y_T \}$ to evaluate the ability of our model. Fig.~\ref{fig:male} shows the results for the gender attribute(female $\to$ male). Our model can obtain a consistent male appearance while accurately synthesizing the hands and face, encoding the non-manual features. Our experiments show that the encoder-decoder network $\mathbf{G}$ does not need to implement temporal consistency since it is already encoded in the input $\mathcal{Y}$.

\begin{figure*}[h!]
    \centering
    \includegraphics[width=.8\textwidth]{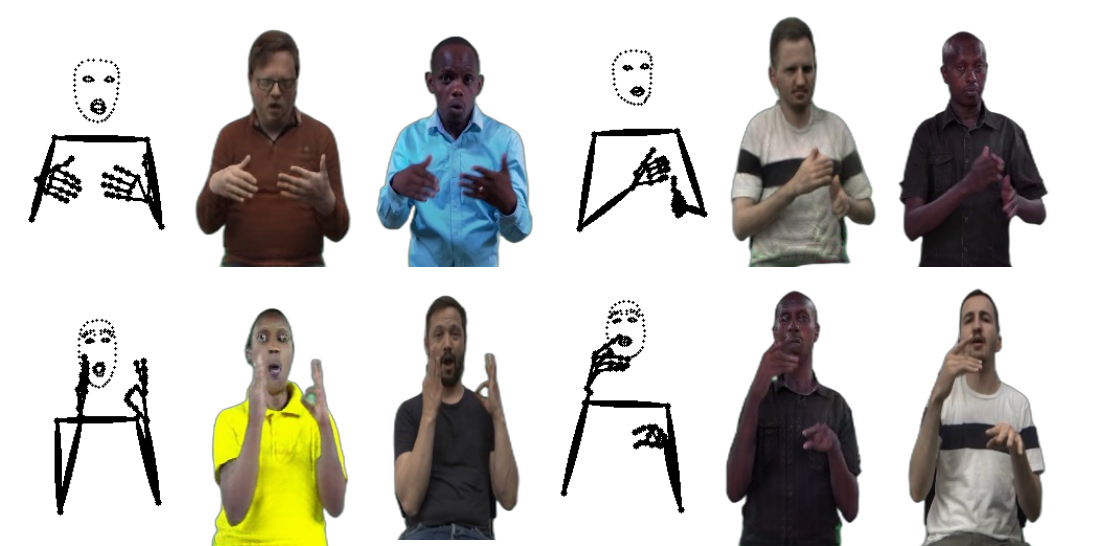}
    \caption{\textbf{Skin synthesis}. Examples showing the synthesis of skin color signers. KEY -- top-row: black $\to$ white signer; bottom-row: white $\to$ black signer.}
    \label{fig:soa_skin}
\end{figure*}

\textbf{Attribute synthesis} It is worth noting that our model can faithfully represent non-manual features of the target attribute. For example, in Fig.~\ref{fig:soa_skin}, our model can synthesize the open mouth (left part) and the head down (top-left). PENet can even generate images with self-occlusion (bottom row). Our model tends to neglect motion blur due to frame-rate sampling. For example, as shown in Fig.~\ref{fig:soa_ethnicity} (top-left), the synthesised Swiss signer contains no blur, unlike the Japanese signer.

PENet learns a fine-grained representation of body parts without explicit priors. For example, hairstyle is blended with clothing style, while the relevant information from the conditioned attribute remains fixed. It is also crucial for the skin attribute that the skin tone on the arm and face of the synthesised images is uniform.

\begin{figure}[h!]
    \centering
    \includegraphics[width=.35\textwidth]{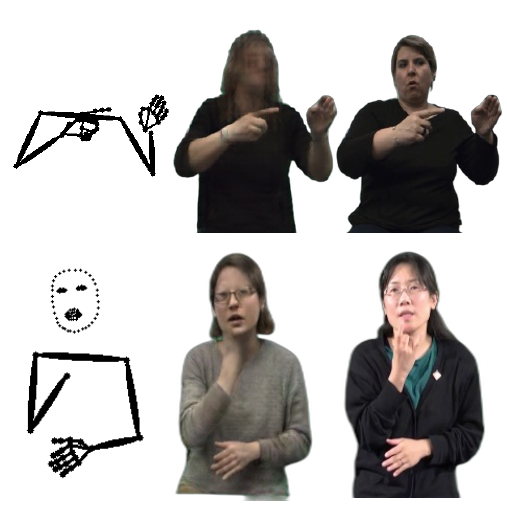}
    \caption{\textbf{Failure cases}. We note that partial skeleton estimation leads to failure in the synthesis. KEY -- left: input pose; middle: synthesised signer image; right: ground-truth signer image.}
    \label{fig:fail}
\end{figure}

\textbf{Challenging cases.} Despite the excellent performance of our proposed model, we found some challenging cases. In particular, the influence of the quality of the input pose sequence $\mathcal{Y}$. Indeed, partial or wrong pose estimation leads to a wrong synthesis. As shown in Fig.~\ref{fig:fail}, the model cannot cope with partial or missing body parts. Without the head keypoints, the model cannot hallucinate the face (Fig.~\ref{fig:fail} top). Partial arm estimation also leads to partial synthesis, such as a chopped-off hand (Fig.~\ref{fig:fail} bottom).

\begin{figure}[h!]
    \centering
    \includegraphics[width=.5\textwidth]{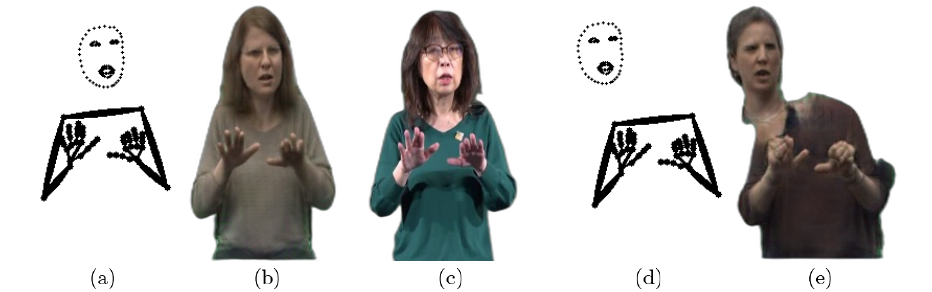}
    \caption{\textbf{Pose conditioning}. We highlight the dependence of our model on pose conditioning. In this example, we change the head landmarks position to further left and see that. Indeed, the head is generated according to the updated landmark. KEY -- (a): conditioning pose $y$; (b): synthesised image; (c): ground-truth image; (d): modified pose; (e): synthesised image with the modified pose.}
    \label{fig:wrongpose}
\end{figure}

However, we note that the model is robust to the keypoint position. This is a suitable property since correcting the pose sequence $\mathcal{Y}$ can improve synthesis without further model modification. To verify this observation, we perform the following experiment. We change the position of a person's landmarks and see that the model synthesizes the signer with the updated keypoint position. For example, in Fig.~\ref{fig:wrongpose}, we change the position of the head and can verify that the model generates an image with the changed position of the head.

\section{Conclusions} \label{sec:conc}
This paper addresses the problem of diversity-aware Sign Language production. In particular, we address sampling diversity within the same category class and a set of attributes defined in our work, such as gender, skin color, and ethnicity.
To achieve this, we propose a GAN-based encoder-decoder framework where we efficiently incorporate the 2D pose conditioning into the variational inference during the training phase. In particular, we use a UNet model as a base generator. The goal is to retain the spatial information from the 2D pose and use visual features from the VAE branch to synthesize an image of a person with the given pose. 

The proposed Pose Encoding VAE learns a better agnostic feature representation of signers. Additionally, we can manipulate the attribute to synthesize diverse signers in the same pose. Moreover, our proposed edge loss adds high-frequency details, resulting in better visual quality of images. We empirically demonstrated that our proposed PE-VAE significantly improves the visual quality of the synthesised images.
Extensive experimental evaluations validate the proposed PENet against strong baseline models.


\blfootnote{ \textbf{Acknowledgement.} This work was supported by the SNSF project `SMILE II' (CRSII5 193686), European Union's Horizon2020 programme (`EASIER' grant agreement 101016982) and the Innosuisse IICT Flagship (PFFS-21-47). This work reflects only the authors view and the Commission is not responsible for any use that may be made of the information it contains.}


{\small
\bibliographystyle{ieee}
\bibliography{refs}
}

\end{document}